\documentclass[runningheads]{llncs}
\usepackage[T1]{fontenc}
\usepackage{graphicx}
\usepackage{caption}
\usepackage{subcaption}
\usepackage{listings}
\usepackage{xcolor}

\usepackage{tikz}
\usetikzlibrary{arrows.meta,positioning,fit,calc,backgrounds,shapes.multipart,shapes.symbols,shapes.geometric}

\tikzset{
  stage/.style={
    rectangle, rounded corners=2pt, draw=black, very thick, align=left,
    inner sep=6pt, fill=white, text width=4.8cm
  },
  data/.style={
    cylinder, shape border rotate=90, aspect=0.25,
    draw=black, very thick,
    minimum height=13mm, minimum width=16mm,
    fill=white, cylinder body fill=white, cylinder end fill=white
  },
  callout/.style={
    rectangle, draw=black, very thick,
    rounded corners=2pt, align=left, inner sep=4pt,
    fill=white, text width=6.3cm
  },
  outbox/.style={
    rectangle, rounded corners=2pt, draw=black, very thick,
    align=left, inner sep=6pt, fill=white, text width=4.8cm
  },
  >={Latex[length=2.2mm]},
  flow/.style={-Latex, very thick, draw=black},
  ref/.style={dashed, -Latex, thick, draw=black},
  title/.style={font=\bfseries},
}

\usepackage{float}
\usepackage{booktabs}
\usepackage{multirow}
\usepackage{todonotes}
\usepackage{amsmath}
\usepackage{enumitem}
\begin{document}
\title{Competency Questions as Executable Plans: a Controlled RAG Architecture for Cultural Heritage Storytelling}
\titlerunning{A CQ-Driven RAG Workflow for Digital Storytelling}
%
%
\author{Naga Sowjanya Barla\orcidID{0009-0001-2691-7672} \and
\\Jacopo de Berardinis\orcidID{0000-0001-6770-1969}}

\authorrunning{N.S. Barla and J. de Berardinis}
%
\institute{School of Computer Science and Informatics, University of Liverpool,\\Brownlow Hill, Liverpool, L69 7ZX, United Kingdom \\
\email{\{n.barla,jacodb\}@liverpool.ac.uk}}
\maketitle

\begin{abstract}
The preservation of intangible cultural heritage is a critical challenge as collective memory fades over time.
While Large Language Models (LLMs) offer a promising avenue for generating engaging narratives, their propensity for factual inaccuracies or ``hallucinations'' makes them unreliable for heritage applications where veracity is a central requirement.
To address this, we propose a novel neuro-symbolic architecture grounded in Knowledge Graphs (KGs) that establishes a transparent ``plan-retrieve-generate'' workflow for story generation.
A key novelty of our approach is the repurposing of competency questions (CQs) -- traditionally design-time validation artifacts -- into run-time executable narrative plans.
This approach bridges the gap between high-level user personas and atomic knowledge retrieval, ensuring that generation is evidence-closed and fully auditable.
We validate this architecture using a new resource: the \textbf{Live Aid KG}, a multimodal dataset aligning 1985 concert data with the Music Meta Ontology and linking to external multimedia assets.
We present a systematic comparative evaluation of three distinct Retrieval-Augmented Generation (RAG) strategies over this graph: a purely symbolic \emph{KG-RAG}, a text-enriched \emph{Hybrid-RAG}, and a structure-aware \emph{Graph-RAG}.
Our experiments reveal a quantifiable trade-off between the factual precision of symbolic retrieval, the contextual richness of hybrid methods, and the narrative coherence of graph-based traversal.
Our findings offer actionable insights for designing personalised and controllable storytelling systems.

\end{abstract}

\section{Introduction}

Intangible cultural heritage -- the ephemeral yet foundational fabric of collective experience -- is inherently fragile  \cite{unesco2003convention}.
Landmark socio-cultural events, such as the 1985 Live Aid concert, represent a complex interplay of tangible artefacts, like the stages at Wembley and JFK Stadiums, and a rich tapestry of intangible elements: iconic musical performances, a unique global broadcast, and a powerful message of humanitarianism.
As direct human memory of such moments recedes, society increasingly relies on digital systems to preserve and transmit this cultural legacy.
Digital storytelling has emerged as a critical medium for this purpose, capable of weaving together disparate facts, media, and contexts into engaging narratives that transform passive observers into active participants in history \cite{vayanou2012towards}.
The evolution of this field in cultural heritage contexts has moved beyond static digital archives towards interactive, personalised experiences that can create ``digital biographies'' of objects and events \cite{huang2023using}.

The advent of Large Language Models (LLMs) has offered unprecedented capabilities for creating fluent and compelling narratives \cite{brown2020gpt3}.
However, this technological leap presents a significant dilemma for applications in cultural heritage.
LLMs are prone to factual invention and distortion -- a phenomenon widely known as ``hallucination'' -- which renders them fundamentally unsuitable for domains where factuality is paramount \cite{ji2023survey,zhang2025survey}.
For museums, archives, and educational platforms, factual veracity is not merely a desirable feature but an ethical and institutional imperative \cite{vayanou2012towards,zhu2025knowledge}.
The core of this issue is a deep epistemological conflict.
LLMs are probabilistic systems; their mastery lies in statistical correlation and linguistic fluency, not in the representation of factual, verifiable knowledge \cite{hogan2021knowledge}.
In contrast, cultural heritage is a domain of evidence, where the value of a statement is intrinsically tied to its provenance \cite{edmond2020digital}.

Retrieval-Augmented Generation (RAG) has become the referential paradigm for mitigating hallucination by conditioning an LLM's output on externally retrieved information \cite{lewis2020retrieval}.
While this represents a significant advance, the retrieval of ambiguous, context-poor, or conflicting text chunks can still lead to fragmented or subtly inaccurate narratives \cite{peng2024graph}.
On the other hand, Knowledge Graphs (KGs) offer a robust foundation for grounding, providing a semantic backbone of unambiguous entities (identified by URIs), explicit, typed relationships, and a formal structure that supports verifiable querying \cite{hogan2021knowledge,zhu2025knowledge}.
KGs ensure that the evidence provided to the generator is atomic, accurate, and traceable, forming the basis for truly dependable storytelling \cite{de2024nodes}.

While KGs provide the verifiable \textit{what}, a research gap remains in defining a methodology for controlling the \textit{how} in this domain.
There is currently no established neuro-symbolic methodology for translating high-level user intent, such as the information needs of a museum visitor, into a structured, auditable, and machine-executable narrative plan that leverages this KG backbone.

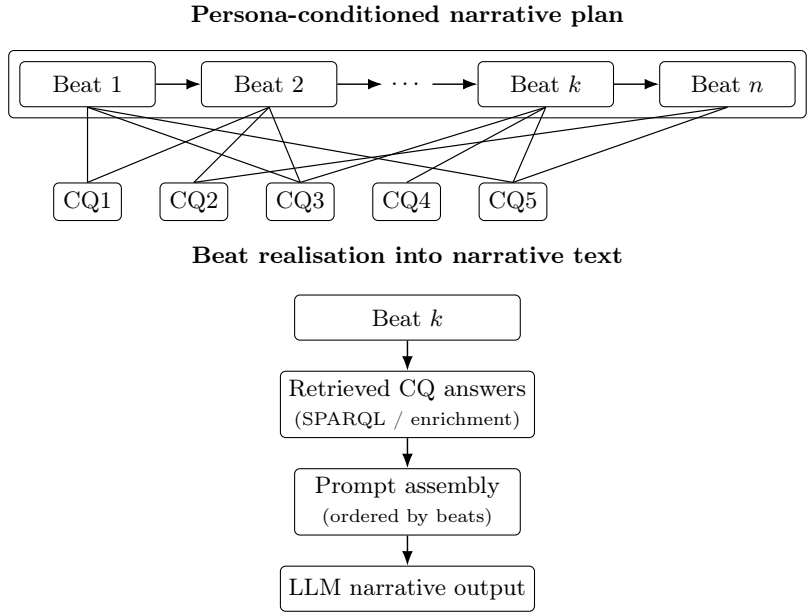
\begin{figure}[t]
\centering
\captionsetup{font=footnotesize}

\begin{subfigure}{\linewidth}
\centering
\begin{tikzpicture}[
  font=\footnotesize,
  beat/.style={draw, rounded corners=2pt, minimum width=18mm, minimum height=6mm, align=center},
  cq/.style={draw, rounded corners=2pt, minimum width=9mm, minimum height=5mm, align=center},
  link/.style={-Latex, line width=0.55pt},     
  soft/.style={line width=0.45pt}             
]


\node[beat] (b1) {Beat 1};
\node[beat, right=6mm of b1] (b2) {Beat 2};
\node[right=6mm of b2] (b3) {$\dots$};
\node[beat, right=6mm of b3] (bk) {Beat $k$};
\node[beat, right=6mm of bk] (bn) {Beat $n$};

\draw[link] (b1) -- (b2);
\draw[link] (b2) -- (b3);
\draw[link] (b3) -- (bk);
\draw[link] (bk) -- (bn);

\node[
  draw,
  rounded corners=2pt,
  inner sep=4pt,
  fit=(b1)(b2)(b3)(bk)(bn),
  label={[yshift=2mm]\textbf{Persona-conditioned narrative plan}}
] {};

\node[cq, below=10mm of b1] (cq1) {CQ1};
\node[cq, right=5mm of cq1] (cq2) {CQ2};
\node[cq, right=5mm of cq2] (cq3) {CQ3};
\node[cq, right=5mm of cq3] (cq4) {CQ4};
\node[cq, right=5mm of cq4] (cq5) {CQ5};

\draw[soft] (b1.south) -- (cq1.north);
\draw[soft] (b1.south) -- (cq3.north);
\draw[soft] (b1.south) -- (cq5.north);

\draw[soft] (b2.south) -- (cq1.north);
\draw[soft] (b2.south) -- (cq2.north);
\draw[soft] (b2.south) -- (cq3.north);

\draw[soft] (bk.south) -- (cq3.north);
\draw[soft] (bk.south) -- (cq4.north);
\draw[soft] (bk.south) -- (cq5.north);

\draw[soft] (bn.south) -- (cq2.north);
\draw[soft] (bn.south) -- (cq5.north);

\end{tikzpicture}
\end{subfigure}

\vspace{2mm}

\begin{subfigure}{\linewidth}
\centering
\begin{tikzpicture}[
  font=\footnotesize,
  box/.style={draw, rounded corners=2pt, minimum width=30mm, minimum height=6mm, align=center},
  link/.style={-Latex, line width=0.55pt},
  node distance=4mm
]
\node[box] (beat) {Beat $k$};
\node[box, below=of beat] (retr) {Retrieved CQ answers\\\scriptsize (SPARQL / enrichment)};
\node[box, below=of retr] (prompt) {Prompt assembly\\\scriptsize (ordered by beats)};
\node[box, below=of prompt] (llm) {LLM narrative output};

\draw[link] (beat) -- (retr);
\draw[link] (retr) -- (prompt);
\draw[link] (prompt) -- (llm);

\node[above=3mm of beat, align=center] {\textbf{Beat realisation into narrative text}};
\end{tikzpicture}
\end{subfigure}

\caption{Story construction as a list of beats based on a shared CQ pool. Each beat is realised by retrieving CQ answers and prompting an LLM in beat order.}
\label{fig:story-construction}
\end{figure}

This paper introduces a novel RAG-based solution by repurposing competency questions (CQs) -- natural language questions that an ontology should satisfy \cite{gruninger1995methodology}.
Traditionally, CQs are a design-time artefact used in various ontology engineering methodologies \cite{presutti2009extreme,suarez2011neon} to define the scope of a knowledge base and to validate its completeness \cite{bezerra2023use}.
We investigate their dual use as requirement elicitation paradigm for knowledge graph construction (the ``\textit{what}'') and as a \textbf{run-time primitive} to orchestrate the entire generative workflow (the ``\textit{how}'').
In our approach, a narrative is conceptualised as a ``beat plan'' -- an ordered sequence of CQs -- that explicitly dictates the information-seeking steps required to construct the story, as illustrated in Fig.~\ref{fig:story-construction}.
This transforms the creative act of storytelling into a transparent, reproducible, and auditable process.

The contributions of this paper are threefold:
\begin{enumerate}
    \item \textbf{The Live Aid KG:} The first publicly available semantic resource dedicated to the 1985 Live Aid concert, modelled using general and domain-specific semantic models  (\texttt{schema.org}, Music Meta Ontology \cite{deberardinis2023music}) and including multimodal links to external assets to ensure interoperability and richness.
    \item \textbf{A CQ-driven RAG workflow:} A novel ``plan-retrieve-generate'' neuro-symbolic pipeline where a sequence of CQs acts as an executable narrative plan, ensuring that every generated sentence is grounded in evidence retrieved from the KG with full provenance.
    \item \textbf{A comparative evaluation:} An empirical analysis of three distinct KG-based RAG strategies -- symbolic \emph{KG-RAG}, text-enriched \emph{Hybrid-RAG}, and structure-aware \emph{Graph-RAG} -- providing actionable insights into the trade-offs between factual precision, contextual richness, and narrative coherence.
\end{enumerate}

Our work aims to empower institutions to valorise semantic resources for visitor engagement, acting as an incentive for the publication of Linked Data.

The paper is organised as follows: Section~\ref{sec:related} reviews related work; Section~\ref{sec:liveaid-kg} describes the Live Aid KG; Section~\ref{sec:cq-rag} details the storytelling workflow; and Sections~\ref{sec:experiments} and \ref{sec:discussion} present the comparative evaluation and discuss our findings.


\section{Background and related work}\label{sec:related}
The use of technology to convey narratives in cultural heritage contexts has a rich history, evolving from early hypermedia systems to modern AI-driven platforms \cite{peng2024graph}.
The goal is often to create personalised and interactive experiences that move beyond static exhibits, allowing users to actively engage with historical artefacts and stories  \cite{nikolakopoulou2025framework}.
KGs have become increasingly central to this effort, providing a formal structure to represent the complex web of entities and relationships inherent in cultural domains \cite{vanerp2023ode,deberardinis2023polifonia}.
However, a persistent challenge remains in bridging the gap between this structured, symbolic knowledge and the generation of fluid and personalised natural language narratives \cite{dong2022survey}.

Recent research has explored using KGs to construct ``narrative graphs'' or ``storylines'', which represent sequences of events and their connections \cite{blin2022building}.
These approaches acknowledge the potential of KGs as a source for narrative content.
Furthermore, a key driver in museum-focused applications is personalisation, where systems adapt content to specific visitor profiles, interests, or knowledge levels to create more meaningful experiences \cite{vayanou2012towards}.
While these works establish the value of KGs for structuring heritage data and personalising content, the challenge of generating accurate and coherent narratives from structured knowledge bases remains an active area of research.

Retrieval-Augmented Generation (RAG) was introduced as a general-purpose approach to equip LLMs with knowledge from external sources, thereby reducing factual invention and allowing knowledge to be updated without retraining the model \cite{lewis2020retrieval}.
The paradigm has since evolved from its initial ``naïve'' implementation, which simply retrieves text chunks and prepends them to a prompt, to more advanced and modular architectures that incorporate techniques for query transformation, document re-ranking, and context compression \cite{peng2024graph}.
Despite these advances, the effectiveness of any RAG system is fundamentally constrained by the quality and structure of the retrieved information.
Retrieving flat, unstructured text chunks can often lead to fragmented or incomplete context, motivating the exploration of more structured knowledge sources to provide better-grounded and more coherent evidence for the generator \cite{lewis2020retrieval,li2021fewshotkg2text}.

Furthermore, the evaluation of RAG systems is itself a complex and growing field, with a suite of metrics developed to assess distinct aspects of performance, including: the relevance of retrieved context (\textit{context relevance}), the factual alignment of the output to the context (\textit{faithfulness} or \textit{answer correctness}), and the degree to which the context contains sufficient information (\textit{context sufficiency}) \cite{min2023factscore}.
The evaluation metrics in our experiments are inspired by this line of work, focusing on fine-grained, evidence-based verification of generated claims.

\textbf{Knowledge graph-augmented generation}.
To address the limitations of unstructured text retrieval, a growing body of work has focused on augmenting RAG pipelines with KGs.
This research has largely bifurcated into two distinct paradigms.
The first, which we term \textbf{KG-RAG}, treats the KG as a structured database.
It uses formal query languages like SPARQL to retrieve precise, atomic facts (i.e., triples) in response to a query.
The second paradigm, known as \textbf{Graph-RAG}, treats the KG as a rich contextual map.
Instead of retrieving isolated facts, it extracts a subgraph or a neighbourhood of nodes and edges around seed entities identified in the query \cite{peng2024graph}.
This subgraph, often linearised into text, provides the LLM with a more holistic view of the local relational context, which can significantly improve the coherence and flow of the generated text \cite{peng2024graph}.
While several works have explored one of these approaches, our work is among the first to implement and systematically compare KG-RAG, Graph-RAG, and a Hybrid variant within a single, unified framework, providing empirical evidence of their distinct performance characteristics for storytelling.

\textbf{Competency questions in knowledge engineering}.
Competency Questions (CQs) are a foundational tool in ontology engineering, serving as natural language questions that an ontology must be able to answer \cite{gruninger1995methodology}.
Their primary role has been to define the scope of a knowledge representation and to serve as a validation suite to ensure the resulting KG meets its design requirements \cite{blomqvist2012ontology}.
While the use of CQs for building and validating ontologies and KGs has been extensively studied \cite{presutti2009extreme}, their potential as a core component in using KGs for generative tasks has been largely unexplored.

Our work introduces a novel application of CQs, repositioning them from a design-time artefact to a run-time driver for narrative generation.
The process of translating abstract user needs (personas and user stories) into a structured sequence of CQs formalises the creative act of storytelling into a series of verifiable information-seeking steps.
\section{The Live Aid Knowledge Graph}\label{sec:liveaid-kg}

The foundational component of the workflow is a bespoke, validated Knowledge Graph (KG) to represent the 1985 Live Aid concert.
The KG serves as the verifiable ``ground truth'' for all narrative generation, making its design, scope, and quality paramount.
This section details its design principles, the multi-stage curation and enrichment process, and its final profile.

\subsection{Requirement collection}

The requirements for the Live Aid Knowledge Graph (KG) were collected through a structured, persona-driven methodology combining narrative analysis, competency question engineering, and semi-automated refinement \cite{deberardinis2023polifonia}.

Live Aid was a landmark dual-venue concert held in 1985 to raise funds for the Ethiopian famine, widely regarded as one of the most significant cultural moments of the 20th century due to its unprecedented global broadcast and humanitarian impact.
To celebrate the 40th anniversary of this historic event, the British Music Experience (BME)\footnote{\texttt{\url{https://www.britishmusicexperience.com}}}, located in the UNESCO City of Music, Liverpool, hosted a temporary, dedicated exhibition\footnote{\texttt{\url{https://www.britishmusicexperience.com/liveaid40}}}.

The museum's curatorial team was involved in the requirement collection phase to ensure the system could support visitor engagement suitable for this application.
Consequently, the project defined two personas that reflect expected visitor types for the exhibition: \textit{Luca}, an informed music enthusiast already present in BME’s previous design work; and \textit{Emma}, a newly introduced novice persona representing first-time visitors with no prior knowledge of Live Aid. These personas were curated with the BME to capture distinct informational needs and to ensure that the KG would support both contextual onboarding and deeper curatorial exploration.

\begin{table}[t]
\centering
\caption{Profile of the Live Aid Knowledge Graph.}\label{tab:kg_profile}
\begin{tabular}{@{}lr@{}}
\toprule
\textbf{KG property} & \textbf{Value} \\
\midrule
Total Triples & 20,343 \\
Distinct Classes & 40 \\
Distinct Predicates (excl. rdf:type) & 109 \\
Typed Subjects & 3,547 \\
Subjects with Dual Typing & 398 \\
\bottomrule
\end{tabular}
\end{table}

Each persona was formalised into a user story describing how they would encounter, interpret, and navigate a Live Aid narrative \cite{carriero2019arco}.
These stories provided the initial substrate from which data requirements were extracted through competency questions.
For example, Emma's story generated CQs relating to foundational event metadata (dates, venues, global reach, humanitarian motivation), whereas Luca's story produced CQs concerning performance sequencing, archival provenance, broadcast coordination, and cross-venue temporal alignment.

To support completeness and consistency, we applied a semi-automated CQ extraction workflow using an LLM.
The model was prompted with persona profiles and user stories to produce expanded lists of candidate CQs, including implicit and fine-grained information needs that might be overlooked in manual decomposition.
These outputs were subsequently deduplicated, scoped, and normalised as done in \cite{deberardinis2023polifonia}, ensuring that each CQ was atomic, linguistically consistent, and answerable using structured data.
The process was supported by automation scripts that generated templated prompts, logged intermediate outputs, and produced structured CQ tables for traceability.

The resulting CQ set constitutes the formal requirements for the Live Aid KG.
Each CQ was mapped to (i) the data it requires, (ii) the ontology fragments needed to express it, and (iii) the external sources from which the corresponding facts would be harvested (e.g. MusicBrainz\footnote{\texttt{\url{https://musicbrainz.org}}} for performance metadata, Wikipedia for contextual and historical information, Live Aid setlist archives for sequencing, and archival notes for broadcast preservation).
This pipeline ensures that KG construction is directly driven by documented user needs and remains aligned with the cultural, narrative, and archival dimensions of the Live Aid.

\subsection{From requirements to ontology reuse}

Driven by the requirements identified before, the KG was designed to reuse existing ontologies, schemas, and controlled vocabularies, in order to promote its interoperability and reuse.
Precisely, we directly reused the following resources.

\begin{itemize}
    \item \textbf{Schema.org}: used for general, high-level concepts such as events (\texttt{schema:Event}), 
    places (\texttt{schema:Place}), and organisations (\texttt{schema:Organization}). Its widespread use ensures broad compatibility with web-based tools and search engines \cite{guha2016schema}.

    \item \textbf{The Music Meta Ontology (\texttt{mm})}: a domain ontology used to represent music metadata, such as musical works (\texttt{mm:MusicEntity}), 
    performances (\texttt{mm:Performance}), and artists (\texttt{mm:MusicArtist}) \cite{deberardinis2023music}.

    \item \textbf{Polifonia Core Ontology (\texttt{core})}: Provides lightweight abstractions for agents, activities, their roles and contextual relationships, enabling alignment with broader European music-heritage resources and supporting interoperability with Polifonia's KGs \cite{deberardinis2023polifonia}.

    \item \textbf{DCMI Metadata Terms (\texttt{dct})}: reused for generic descriptive metadata, including titles, descriptions, dates, creators, and provenance statements. These predicates ensure compatibility with digital-heritage metadata practices and aims to improve long-term maintainability.

    \item \textbf{Web Annotation Data Model (\texttt{oa})}: applied to semantically describe curatorial notes, narrative fragments, and explanatory comments attached to specific entities or facts, enabling fine-grained annotation of Web documents within the KG (in addition to linking them to relevant entities).

    \item \textbf{SKOS (Simple Knowledge Organization System) (\texttt{skos})}: utilised to encode controlled vocabularies such as genres, thematic categories, and narrative labels, supporting hierarchical and associative relationships to improve consistency in classification and retrieval.
\end{itemize}

The direct reuse of existing semantic models ensures that the KG can be easily understood, integrated with other semantic datasets, and queried by various tools and stakeholders.
A key pragmatic design choice was the use of \textbf{dual typing} for artist entities. For example, the band Queen is represented as an instance of both \texttt{schema:MusicGroup} and \texttt{mm:MusicEnsemble}.
This intentional redundancy simplifies the authoring of SPARQL queries by allowing a single, flexible template to retrieve artist information regardless of which vocabulary is being primarily targeted in a given query, thereby improving the reusability and robustness of our CQ templates.

\subsection{Knowledge graph population and enrichment}\label{ssec:kg-enrichment}
The KG was populated through a semi-automated, multi-stage process designed to ensure both breadth of coverage and data quality.

The process began with two canonical seed identifiers from MusicBrainz, representing the Live Aid concerts in London and Philadelphia. From these anchors, we programmatically expanded the graph by fetching and integrating data from open knowledge sources, primarily MusicBrainz and Wikidata. This automated enrichment included the following key steps:
\begin{itemize}
    \item \textbf{Performers and participation:} event line-ups were ingested to link performers to their respective event and venue using \texttt{schema:performer} and \texttt{schema:location}. Each act was typed as either a music group (\texttt{mm:MusicEnsemble}) or a solo artist (\texttt{mm:MusicArtist}).
    \item \textbf{Setlists and timings:} official setlists from the Live Aid release were parsed to create distinct instances for each performance (\texttt{mm:Performance}), linking them to the musical works performed (\texttt{mm:MusicalWork}) and recording their start and end times using \texttt{schema:startDate} and \texttt{schema:endDate}.
    \item \textbf{Identity and interlinking:} entities were enriched with \texttt{owl:sameAs} links to their Wikidata and MusicBrainz counterparts to foster interoperability. External identifiers for other relevant sources, such as Genius\footnote{\texttt{\url{https://genius.com}}} and Songfacts\footnote{\texttt{\url{https://www.songfacts.com}}}, were resolved via Wikidata and added as further contextual hooks.
    \item \textbf{Multimedia links:} where available, links to stable images (\texttt{schema:image}) and publicly available performance videos (\texttt{schema:contentUrl}) were added to support multimedia applications and future extensions of this work.
\end{itemize}

\subsubsection{Curated enrichment for hybrid-RAG}
For the Hybrid-RAG variant, a separate manual curation step was performed to inject richer, more descriptive context. We identified a small set of high-quality web sources, such as retrospective articles from the BBC and CNN, and created short, vetted textual summaries related to key entities and events (e.g., the legacy of Queen's performance). These summaries were modelled as annotations (e.g., using \texttt{skos:note}) and stored locally with full provenance, including the source URL and access date. This controlled process allows the Hybrid-RAG pipeline to supplement the KG's atomic facts with richer narrative context without introducing the potential noise and unreliability of live Web searches \cite{yoran2023making}.


The resulting knowledge graph provides a rich, interconnected representation of the Live Aid event.
Table \ref{tab:kg_profile} presents a summary of its key statistics.
The complete KG is serialised in Turtle format and is made publicly available under Creative Commons Attribution-NonCommercial 4.0 International (CC-BY-NC) licence at \texttt{\url{https://github.com/KE-UniLiv/cq-storyrag}}.

\subsection{Validation, profile, and availability}

To ensure the quality and reliability of the KG, an automated validation process was integrated into the release pipeline. Before any version of the KG is used for experiments, it must pass a suite of checks, including: (1) \textbf{schema conformance}, to ensure all triples adhere to the defined vocabularies and no undeclared or out-of-schema terms appear; (2) \textbf{identifier integrity}, to verify consistent URI patterns, prevent duplication, and consolidate equivalence links across modules; and (3) \textbf{CQ coverage Tests}, where the full suite of parameterised SPARQL queries is executed against the graph. 
A new version is only successful if all CQs can be answered and all consistency checks pass, guaranteeing that the KG is complete, narratively ready and fit for storytelling.


\section{CQ-RAG: Competency question-driven storytelling}\label{sec:cq-rag}

This section introduces an end-to-end workflow for generating personalised and factually grounded narratives.
The architecture is designed around the idea of using CQs as the primary control mechanism, ensuring traceability from high-level user intent to the final generated text.
The system follows a modular: $plan \rightarrow retrieve \rightarrow generate$ pipeline, where each stage is fully auditable.

The architecture is implemented as a modular pipeline, with an example illustrated in Figure~\ref{fig:kg-rag-example}.
This design separates the core tasks of narrative planning, evidence retrieval, and text generation into three different modules.

\begin{enumerate}
    \item \textbf{The Planner} translates a high-level user profile and narrative requirements into a structured, machine-executable narrative plan. This plan takes the form of an ordered sequence of competency questions.
    \item \textbf{The Retriever} executes this plan, taking one CQ at a time and gathering a corresponding ``evidence pack'' from the Live Aid KG. We implement three distinct retrieval strategies to explore different trade-offs, all sharing a common input/output schema (c.f. Section~\ref{ssec:retrieval}).
    \item \textbf{The Generator} synthesises the retrieved evidence into a coherent narrative beat. It operates under a strict evidence-closed constraint, ensuring every claim is grounded in the retrieved facts and that full provenance is kept.
\end{enumerate}

This separation ensures that the creative act of storytelling is formalised into a series of transparent, verifiable information-seeking steps.

\subsection{From personas to narrative plans (Planner)}\label{ssec:planner}
The planning stage is where high-level narrative goals are translated into a machine-executable format.
The process is designed to be transparent and reproducible, flowing from abstract user archetypes to a sequence of retrieval actions.

\begin{itemize}
    \item \textbf{Personas and user stories:} the process begins with the persona definition passed as input, such as ``Emma'', a curious novice, and ``Luca'', an informed enthusiast. These personas define the desired tone, depth, and focus of the story. Each persona is associated with a set of user stories that capture the specific information needs of the user (e.g., ``As a casual fan, I want to know what made Queen's performance so special'').

    \item \textbf{Competency question library:} these user stories are then manually decomposed into a library of atomic, parameterised competency questions, each assigned a stable identifier and a template (e.g., CQ-E24: ``Why is [Artist]'s performance at [Event] considered iconic?''). This library serves as the canonical set of information-seeking goals the system can pursue.
    The workflow focuses on \emph{information} seeking CQs rather than high-level scoping questions \cite{keet2024discerningcharacterisingtypescompetency}.
    Each CQ is associated with a parameterised SPARQL template (see Appendix \ref{ssec:cq-query-templates} for examples), forming an explicit ``API contract'' between the curator’s narrative intent and the execution of the system.

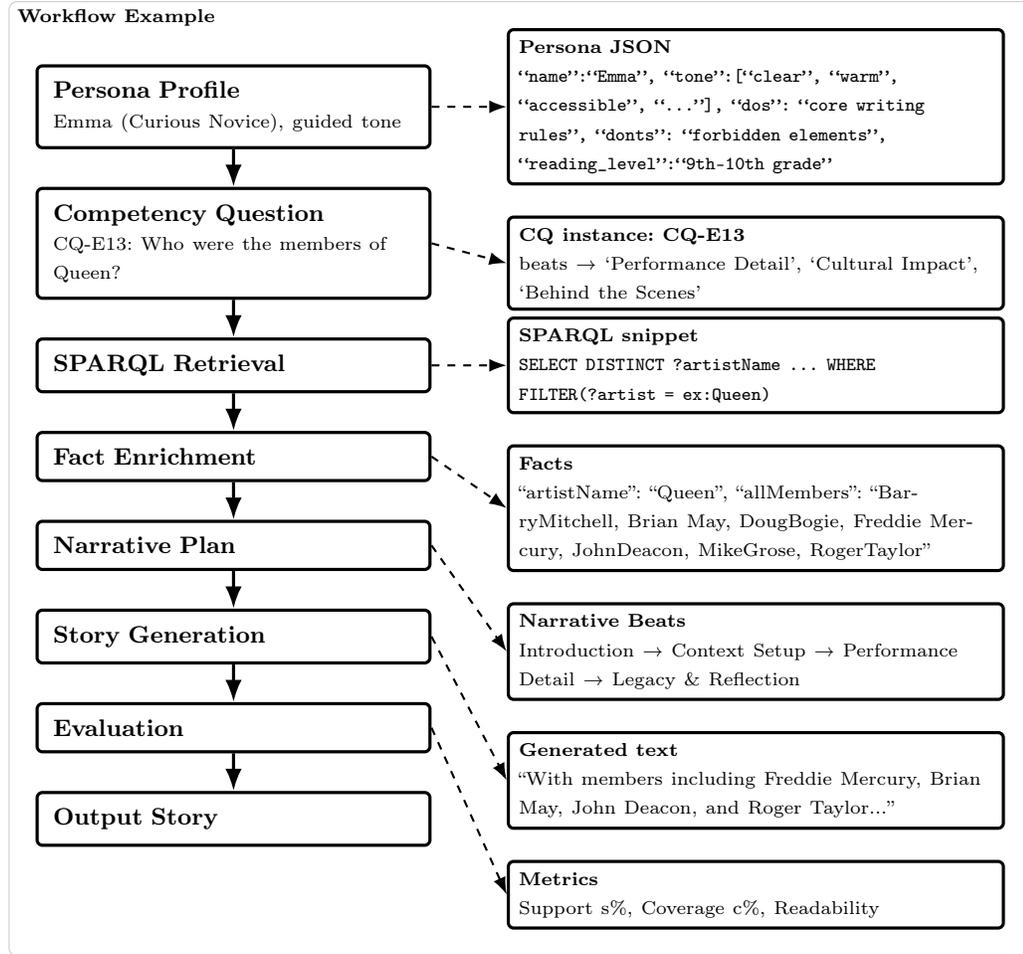
\begin{figure}[H]
\centering
\small
\begin{tikzpicture}[node distance=5mm and 10mm]


\node[stage] (persona) {\textbf{Persona Profile}\\
\scriptsize Emma (Curious Novice), guided tone};

\node[stage, below=of persona] (cqs) {\textbf{Competency Question}\\
\scriptsize CQ-E13: Who were the members of Queen?};
\node[stage, below=of cqs] (sparql) {\textbf{SPARQL Retrieval}};
\node[stage, below=of sparql] (facts) {\textbf{Fact Enrichment}};
\node[stage, below=of facts] (plan) {\textbf{Narrative Plan}};
\node[stage, below=of plan] (gen) {\textbf{Story Generation}};
\node[stage, below=of gen] (eval) {\textbf{Evaluation}};
\node[outbox, below=of eval] (out) {\textbf{Output Story}};

\draw[flow] (persona) -- (cqs);
\draw[flow] (cqs) -- (sparql);
\draw[flow] (sparql) -- (facts);
\draw[flow] (facts) -- (plan);
\draw[flow] (plan) -- (gen);
\draw[flow] (gen) -- (eval);
\draw[flow] (eval) -- (out);

\node[callout, right=10mm of persona] (ex1)
{\textbf{\scriptsize{Persona JSON}}\\
\scriptsize \texttt{``name'':``Emma'', ``tone'':[``clear'', ``warm'', ``accessible'', ``...''], ``dos'': ``core writing rules'', ``donts'': ``forbidden elements'', ``reading\_level'':``9th–10th grade''}};

\node[callout, right=10mm of cqs, below=4mm of ex1] (ex2)
{\textbf{\scriptsize{CQ instance: CQ-E13}}\\
\scriptsize beats → `Performance Detail', `Cultural Impact', `Behind the Scenes'};

\node[callout, right=10mm of sparql] (ex3)
{\textbf{\scriptsize{SPARQL snippet}}\\
\scriptsize \texttt{SELECT DISTINCT ?artistName ... WHERE FILTER(?artist = ex:Queen) }};

\node[callout, right=10mm of facts, below=4mm of ex3] (ex4)
{\textbf{\scriptsize{Facts}}\\
\scriptsize {``artistName'': ``Queen'', ``allMembers'': ``BarryMitchell, Brian May, DougBogie, Freddie Mercury, JohnDeacon, MikeGrose, RogerTaylor''}};

\node[callout, right=10mm of plan, below=4mm of ex4] (ex5)
{\textbf{\scriptsize{Narrative Beats}}\\
\scriptsize Introduction → Context Setup → Performance Detail → Legacy \& Reflection};

\node[callout, right=10mm of gen, below=4mm of ex5] (ex6)
{\textbf{\scriptsize{Generated text}}\\
\scriptsize ``With members including Freddie Mercury, Brian May, John Deacon, and Roger Taylor...''};

\node[callout, right=10mm of eval, below=4mm of ex6] (ex7)
{\textbf{\scriptsize{Metrics}}\\
\scriptsize Support s\%, Coverage c\%, Readability};

\foreach \x/\y in {persona/ex1,cqs/ex2,sparql/ex3,facts/ex4,plan/ex5,gen/ex6,eval/ex7}{
  \draw[ref] (\x.east) -- (\y.west);
}

\node[draw=black!20, rounded corners=3pt,
      fit=(persona)(cqs)(sparql)(facts)(plan)(gen)(eval)(out)(ex1)(ex7),
      inner sep=10pt,
      label={[title,anchor=north west]north west:
      {\scriptsize\textbf{Workflow Example}}}] {};
\end{tikzpicture}
\caption{End-to-end execution of a narrative beat for the \textit{Emma} persona. The diagram illustrates the \textbf{Plan-Retrieve-Generate} cycle. The resulting evidence is synthesized into a narrative segment, grounded in the retrieved facts.}
\label{fig:kg-rag-example}
\end{figure}

    The CQ–SPARQL mappings are manually defined for this study (with LLM-assisted scaffolding) as a one-time setup step to ensure that the narrative plans remain auditable and aligned with the curatorial standards.
    Future iterations will explore leveraging semi-automated CQ-to-SPARQL translation methods (e.g., \cite{emonet2025llmbasedsparqlquerygeneration}).
    Overall, the CQ library serves as the canonical set of information-seeking goals the system can pursue.

    \item \textbf{The beat plan:} the Planner takes a persona and a desired narrative length (\texttt{Small}, \texttt{Medium}, or \texttt{Long}) as input. It constructs a \textbf{beat plan} -- an ordered sequence of these CQs with their parameters bound to specific entities from the KG (e.g., Artist=Queen, Event=Live Aid). To ensure full reproducibility, any random choices in this planning process (e.g., selecting from multiple eligible artists for a beat) are controlled by a fixed seed. The final plan is a JSON object that precisely dictates the narrative's structure and content, ready for execution by the Retriever.
\end{itemize}

\subsection{Competency question infilling (Retrieval strategies)}\label{ssec:retrieval}
The retriever executes the beat plan by processing one CQ at a time and assembling an ``evidence pack'' for the generator.
For each CQ, the system runs the corresponding parameterised SPARQL template from the CQ library; queries are therefore not generated by the LLM at runtime, preventing hallucinated retrieval steps and ensuring institutional control over knowledge access.
We implemented and compared three retrieval strategies with a shared input/output schema, allowing them to be used interchangeably within the pipeline.

\begin{itemize}
    \item \textbf{KG-RAG:} this strategy implements a purely symbolic approach. The CQ's identifier is mapped to a parameterised SPARQL template. The entity bindings from the beat plan (e.g., URIs for `Queen' and `Live Aid') are injected into the template, and the resulting query is executed directly against the in-memory KG. The output is a clean, capped set of RDF triples representing the direct, atomic facts that answer the question. This method prioritises factual precision and clear provenance.

    \item \textbf{Hybrid-RAG:} this strategy combines the precision of SPARQL with the descriptive richness of text. It first performs the KG-RAG step to retrieve core facts. Then, if the CQ is flagged as benefiting from additional context (e.g., questions about legacy or reception), it queries a local, indexed cache of vetted web snippets. These snippets are short, curated summaries from high-quality sources (e.g., BBC, CNN retrospectives). The top-K matching snippets are added to the evidence pack alongside the triples, with their source URL and access date preserved for provenance. This approach enriches the narrative without the noise and latency of live web searches.

    \item \textbf{Graph-RAG:} This approach focuses on retrieving local relational context rather than answering a specific question directly. The entities in the CQ's bindings are used as seeds to expand a small, typed neighbourhood within the KG. The expansion is constrained by a pre-defined relation allowlist and a hop limit (typically 1 or 2 hops) to prevent retrieving irrelevant information. To ensure determinism, any tie-breaking during traversal is controlled by the run's global seed. The resulting subgraph is then flattened into a set of triples and passed to the generator, providing a web of thematically related facts to improve narrative coherence.
\end{itemize}

\subsection{Evidence-closed generation}\label{ssec:generation}
The final stage synthesises the retrieved evidence into a short, fluent narrative beat.
To guarantee factuality and prevent hallucination, the generator (an LLM) operates under a strict \textbf{evidence-closed} constraint: it is explicitly prompted to construct sentences using \textit{only} the facts contained within the provided evidence pack and is forbidden from introducing new entities or claims.

This is implemented as a two-step routine: (i) a \textbf{content pass} selects and orders a minimal set of salient facts from the evidence pack to form a micro-outline for the beat; (ii) a \textbf{surface pass} realises this outline into 1--3 fluent sentences, applying light connectors (from supporting facts to full sentences) and persona-specific stylistic controls (e.g., tone, sentence length).

This process is designed to be fully auditable.
For each generated narrative, the system logs a sentence-to-evidence map, which records precisely which triple(s) or snippet(s) from the evidence pack support each sentence in the output.
As a final safeguard, if an evidence pack is empty or too sparse to generate a meaningful beat, the beat is suppressed entirely, and the reason is logged.
This prevents the generation of ungrounded or potentially hallucinated content.

\section{Experimental evaluation}\label{sec:experiments}

To assess the effectiveness of our workflow and compare the performance of the three RAG variants, we conducted a systematic experimental evaluation.

We generated narratives for both the ``Emma'' (novice) and ``Luca'' (enthusiast) personas across three lengths: Small, Medium, and Long. For each of the 18 configurations (2 personas $\times$ 3 lengths $\times$ 3 RAG variants), we performed multiple runs with different random seeds to ensure the robustness of our findings.
A total of $n=20$ runs were conducted for the Emma persona and $n=5$ for Luca per configuration.
All experiments were run using \texttt{Gemini 2.5 Flash}, with fixed configurations to ensure reproducibility.
In particular, the default generation parameters (temperature, top-$p$, or top-$k$) from the API client were reused.
Examples of generated narratives are provided in Appendix~\ref{sec:appendix}.

\subsection{Evaluation protocol}

In line with previous works (c.f. Section~\ref{sec:related}), the narrative output is evaluated using five complementary metrics that capture factual grounding, coverage of retrieved evidence, readability, and discourse cohesion.

\begin{itemize}[itemsep=5pt]

    \item \textbf{Support.}
    A measure of fine-grained factual grounding inspired by metrics such as FActScore~\cite{min2023factscore}. Support is computed at the \emph{sentence level}: a sentence is considered supported if it can be directly grounded in the retrieved evidence. This metric is mode-aware; the grounding check adapts to the specific representation of the retrieved evidence while consistently measuring the explicit anchoring of generated sentences to the source content.
    
    Formally, for a section comprising sentences $S=\{s_1,\dots,s_m\}$, the support ratio is defined as:
    {\footnotesize
    $$
        \mathrm{support\_ratio}(S)
        =
        \frac{1}{m}
        \sum_{i=1}^{m}
        \mathbf{1}[\mathrm{supported}(s_i)],
    $$
    }
    where the condition $\mathrm{supported}(s_i)$ evaluates to true based on the active evaluation mode:
    \begin{itemize}
        \item \emph{Factlet mode}: $s_i$ shares at least two non-stopword tokens with a factlet, achieving a token-level Jaccard similarity of $\ge 0.22$;
        \item \emph{Triple mode}: $s_i$ co-mentions the subject and object of an evidence triple;
        \item \emph{Fallback mode}: $s_i$ co-mentions at least two mined entities.
    \end{itemize}
    The \texttt{support\_pct\_mean} metric
    is the mean of this ratio across all sections.

    \item \textbf{Coverage.}  
    A complementary factuality measure that quantifies the extent to which the retrieved evidence is utilised by the model. Like support, this metric is mode-aware; depending on the representation of the retrieved evidence, it measures the proportion of source material that is realised in the generated narrative. In the primary (\emph{factlet}) mode, coverage is computed as:
    {\footnotesize
    $$
        \mathrm{coverage}_{\text{factlet}}(S)
        =
        \frac{1}{|F|}
        \sum_{i=1}^{|F|}
        \mathbf{1}\bigl[ |T(S)\cap T(f_i)| \ge 2 \bigr],
    $$
    }
    where $F$ is the set of factlets and $T(\cdot)$ is the content-token function.  
    In \emph{triple mode}, coverage is the fraction of top-degree KG entities that appear in the normalised output.  
    The final value reported is \texttt{coverage\_pct\_mean}.


    \item \textbf{Readability.} 
    This is measured using the Flesch-Kincaid Reading Ease \cite{kincaid1975}:
    {\footnotesize
    \[
        \mathrm{FRE}(S)
        =
        206.835
        - 1.015\!\left(\frac{W}{S_{\text{sent}}}\right)
        - 84.6\!\left(\frac{\mathrm{syllables}}{W}\right),
    \]
    }
    where $W$ is the word count and $S_{\text{sent}}$ the number of sentences.  
    The aggregate score is the mean FRE across sections.

    \item \textbf{Local Cohesion.}
    This metric measures the narrative flow \emph{within a beat}. It is operationalised as the fraction of adjacent sentence pairs whose token-level Jaccard similarity falls within a defined threshold range, ensuring semantic continuity without excessive repetition:
    
    {\footnotesize
    \[
        \mathrm{local\_cohesion}(S)
        =
        \frac{
            |\{\, i \mid 0.15 \le \mathrm{Jaccard}(T(s_i), T(s_{i+1})) \le 0.65 \,\}|
        }{m-1}.
    \]
    }
    This approximates the conceptual intent of semantic flow between adjacent sentences.  
    The reported score is \texttt{local\_cohesion\_mean}.

    \item \textbf{Global Cohesion.}
    While local cohesion captures intra-beat flow, global cohesion measures coherence \emph{across} beats. The score aggregates multiple discourse-level signals that contribute to long-form text coherence: entity transitions \cite{barzilay2008modeling}, entity continuity \cite{guinaudeau2013graph}, bridging relations \cite{clark1975bridging}, temporal ordering \cite{pustejovsky2003timeml}, and referential stability \cite{grosz1995centering}. Specifically, we compute:
    {\footnotesize
    $$
    \begin{aligned}
    \mathrm{global\_cohesion}(S) &= 0.35\,\text{local\_cohesion} + 0.25\,\text{entity\_flow} \\
    &\quad + 0.15\,\text{bridge\_rate} + 0.15\,\text{temporal\_consistency} \\
    &\quad + 0.10\,\text{reference\_stability}
    \end{aligned}
    $$
    }
    The weighting scheme reflects the relative importance of these features in coherence modelling: local entity transitions and continuity receive higher weight, while bridging, temporal consistency, and reference stability introduce complementary global structure.




\end{itemize}

\subsection{Results}

The quantitative results, detailed in Table~\ref{tab:results_factuality} and Table~\ref{tab:results_quality}, reveal distinct performance profiles for each retrieval strategy.
The data highlights a quantifiable trilemma between verifiability, narrative richness, and structural cohesion.
Given the limited sample size of the evaluation corpus, we report our results descriptively to highlight the primary performance trends across the metrics.

\textbf{KG-RAG} establishes itself as the baseline for factual precision.
It consistently achieves the highest \textit{Support} scores across most configurations (e.g., peaking at 90.11\% for Luca-Medium), confirming that restricting generation to atomic triples minimises hallucination and ensures that claims are directly traceable to database entries.
In contrast, \textbf{Hybrid-RAG} performs best in \textit{Coverage}, particularly in longer narratives where it consistently exceeds 90\%.
By supplementing the KG with vetted web snippets, this strategy effectively fills ``narrative gaps'' where the graph is sparse.
While its support scores remain competitive (66--75\%), the slight dip compared to KG-RAG suggests that integrating unstructured text introduces a minor degree of noise, although considerably boosting the comprehensive nature of the story.

The \textit{Readability} (FRE) scores provide insight into the ``texture'' of the generated narratives.
\textbf{KG-RAG} produces the most accessible text (FRE $\sim$60).
Qualitative inspection suggests that this is structural: atomic triples tend to induce simpler, declarative sentence structures.
In contrast, \textbf{Hybrid-RAG} produces the lowest readability scores (FRE $\sim$43--48).
This increased density is a byproduct of its source material; by ingesting real-world journalistic snippets (from BBC, CNN, etc.), the model adopts more complex syntactic structures and vocabulary. This confirms that while Hybrid-RAG is richer, it places a higher cognitive load on the reader.

\textbf{Graph-RAG} presents a notable outlier.
It achieves a maximal \textit{Global Cohesion} score of 1 across all runs.
Unlike the other methods, which retrieve isolated facts or snippets, Graph-RAG feeds the generator a pre-connected subgraph.
This forces the narrative to follow the existing edges of the graph, resulting in seamless entity transitions and a structurally rigorous flow.
However, this comes with a sharp penalty in measured grounding, with Support dropping to $\sim$11--32\%.
Nevertheless, we remark that this disconnect is likely an artefact of the evaluation metric: standard triple-matching algorithms struggle to align the high-level summarisation of a complex subgraph with atomic source triples.

\begin{table}[t]
\centering
\caption{Factual grounding metrics (Support and Coverage, in \%) by persona, length, and RAG variant. Higher is better. KG-RAG demonstrates the highest support, while Hybrid-RAG excels in coverage.}\label{tab:results_factuality}
\resizebox{\textwidth}{!}{%
\begin{tabular}{@{}llcc|cc|cc@{}}
\toprule
& & \multicolumn{2}{c}{\textbf{KG-RAG}} & \multicolumn{2}{c}{\textbf{Hybrid-RAG}} & \multicolumn{2}{c}{\textbf{Graph-RAG}} \\
\cmidrule(l){3-8} 
\textbf{Persona} & \textbf{Length} & \textbf{Support\%} & \textbf{Coverage\%} & \textbf{Support\%} & \textbf{Coverage\%} & \textbf{Support\%} & \textbf{Coverage\%} \\
\midrule
\multirow{3}{*}{Emma} & Small & 71.95 & 83.40 & 69.96 & 89.56 & 22.36 & 14.60 \\
& Medium & 73.59 & 88.57 & 67.79 & 89.58 & 17.15 & 14.78 \\
& Long & 75.65 & 87.88 & 69.64 & 92.17 & 16.26 & 14.72 \\
\midrule
\multirow{3}{*}{Luca} & Small & 74.65 & 90.00 & 75.92 & 89.71 & 22.89 & 13.47 \\
& Medium & 76.21 & 90.11 & 66.44 & 90.06 & 32.64 & 19.05 \\
& Long & 71.18 & 87.06 & 69.30 & 91.74 & 11.51 & 11.10 \\
\bottomrule
\end{tabular}%
}
\end{table}
\begin{table}[t]
\centering
\caption{Readability and Cohesion metrics by persona, narrative length, and RAG variant. FRE denotes the Flesch-Kinkaid Reading Ease (higher is easier). Glob/Loc Cohesion is cosine similarity (higher is more cohesive).}\label{tab:results_quality}
\begin{tabular}{@{}llccc|ccc|ccc@{}}
\toprule
& & \multicolumn{3}{c}{\textbf{KG-RAG}} & \multicolumn{3}{c}{\textbf{Hybrid-RAG}} & \multicolumn{3}{c}{\textbf{Graph-RAG}} \\
\cmidrule(l){3-11} 
\textbf{Persona} & \textbf{Length} & \textbf{FRE} & \textbf{Glob} & \textbf{Loc} & \textbf{FRE} & \textbf{Glob} & \textbf{Loc} & \textbf{FRE} & \textbf{Glob} & \textbf{Loc} \\
\midrule
\multirow{3}{*}{Emma} & Small & 58.28 & 0.497 & 0.410 & 48.61 & 0.524 & 0.477 & 50.10 & 1.000 & 0.448 \\
& Medium & 60.65 & 0.517 & 0.402 & 48.38 & 0.551 & 0.480 & 50.72 & 1.000 & 0.417 \\
& Long & 61.07 & 0.501 & 0.403 & 46.73 & 0.523 & 0.471 & 51.81 & 1.000 & 0.417 \\
\midrule
\multirow{3}{*}{Luca} & Small & 58.27 & 0.482 & 0.406 & 43.43 & 0.555 & 0.500 & 53.03 & 1.000 & 0.432 \\
& Medium & 57.67 & 0.515 & 0.402 & 46.19 & 0.535 & 0.491 & 54.03 & 1.000 & 0.422 \\
& Long & 57.86 & 0.485 & 0.430 & 44.17 & 0.525 & 0.458 & 50.96 & 1.000 & 0.415 \\
\bottomrule
\end{tabular}%
\end{table}
\section{Discussion}\label{sec:discussion}

Our findings map out a clear design space for neuro-symbolic storytelling.
The choice of retrieval strategy is not a matter of finding a single ``best'' method, but rather of aligning the retriever's characteristics with the specific requirements.

\begin{itemize}[leftmargin=0pt,itemsep=0pt,topsep=0pt]
    \renewcommand\labelitemi{}
    \item \textbf{Auditability (KG-RAG).} When the primary requirement is institutional trust, the symbolic KG-RAG approach is the most suitable approach. Its strict query-to-fact alignment ensures that every claim is a direct verbalisation of a sourced entry, minimising liability and maximising provenance.
    
    \item \textbf{Engagement (Hybrid-RAG).} When the goal is to immerse the visitor in a broader historical context (e.g., ``digital biographies''), Hybrid-RAG offers the best tradeoff. It leverages the KG for structure but successfully creates a ``thicker'' narrative by weaving in descriptive text, accepting the trade-off in syntactic complexity observed in the results.
    
    \item \textbf{Exploration (Graph-RAG)} For applications prioritising thematic discovery and fluid reading, Graph-RAG shows more potential. It produces the most cohesively structured text, effectively mimicking a curated ``journey'' through the data. However, until evaluation methods evolve to verify subgraph summarisation, it remains risky for high-stakes factual domains.
\end{itemize}

Beyond the retrieval comparison, this work validates the \emph{CQ-driven beat plan} as an effective semantic scaffold that imposes a logical, auditable structure on generation.
We further discuss specific model behaviours in Appendix~\ref{ssec:behaviours}.


The primary limitation identified is the \textit{evaluation gap} for graph-based generation.
Standard reference-free metrics (like our triple alignment) penalise the high-level summarisation characteristic of Graph-RAG.
Future work will develop path-aware alignment metrics that can verify a sentence against a subgraph path rather than a single triple.
Additionally, while our automated metrics provide a proxy for quality, a human-in-the-loop study is necessary to validate subjective qualities such as tone appropriateness and emotional engagement.

We also acknowledge that our experimental validation focused on a single scenario -- the Live Aid KG.
Nevertheless, because narrative beats and CQ alignment operate at the requirements level, the architecture offers a flexible approach that can be extended to other KGs with typed entities and relations.
Adapting the workflow to other KGs primarily requires the formalisation of domain-specific personas, narrative beats, and their corresponding CQ--SPARQL mappings.

Finally, we posit that the explicit reliance on CQs serves as a methodological bridge that incentivises best practices in knowledge engineering.
As noted by Monfardini et al.~\cite{monfardini2023CQSurvey}, competency questions are often employed during ontology development but rarely published.
Releasing CQ libraries alongside SPARQL queries and RDF dumps allows the very artefacts that guide ontology design to be repurposed for controlled, verifiable narrative generation.
\section{Conclusions}
This paper presented a KG-first, RAG-based workflow for generating personalised and factually grounded digital stories about intangible cultural heritage, using the 1985 Live Aid concert as a case study. We introduced a novel methodology where competency questions, derived from user personas, serve as the central mechanism for planning the narrative structure and driving evidence retrieval. Our contributions include the release of the Live Aid KG, the CQ-driven storytelling pipeline, and a systematic comparison of three RAG variants.
The results demonstrate that different retrieval strategies offer distinct advantages, creating a clear trade-off between factual precision (KG-RAG), contextual richness (Hybrid-RAG), and narrative coherence (Graph-RAG).

Future work will focus on expanding the evaluation framework by developing path-level alignment metrics to more accurately score the factuality of Graph-RAG outputs.
We also plan to conduct a human-in-the-loop study to validate our automated metrics against human judgments of quality and engagement.
Second, we will explore the automation of the pipeline, investigating methods to semi-automatically generate CQs from user stories or directly from the KG structure, reducing the manual authoring burden.

\subsubsection*{Supplemental Material Statement} All data and code to reproduce our experiments are available at
(\texttt{\url{https://github.com/KE-UniLiv/cq-storyrag}}).

\section*{Acknowledgements}
This work was supported by the Engineering and Physical Sciences Research Council (EPSRC) [grant EP/X525741/1] via an Impact Acceleration Award, and the UK Robotics and Autonomous Systems (UK-RAS) Network [grant 277886225].
The authors also thank Oliver Taylor and Elizabeth Koravos from the British Music Experience for their support in requirement gathering.

\bibliographystyle{splncs04}
\bibliography{references}

\appendix 
\appendix
\section{Qualitative examples from KG--RAG}\label{sec:appendix}

This appendix presents selected excerpts from the Luca--Long (KG) narratives across multiple runs, illustrating different behaviours observed during qualitative inspection of the system’s outputs. These examples show how the model integrated retrieved KG facts into generated text; where it produced faithful renderings; where it introduced helpful narrative structure; and where it occasionally deviated or fused information in problematic ways. All fragments are drawn directly from stories produced by the workflow.

\subsection{Faithful behaviour (Beat~2: performance detail)}

\textbf{Retrieved KG facts.}  
The KG entries for this beat include (i) a structured fact stating that Queen performed the song \emph{``Hammer to Fall''} in position~3 during Live Aid~1985, and (ii) five separate review statements from BBC, CNN, Rolling Stone, Wikipedia, and uDiscoverMusic, each describing Queen’s Live Aid set as a show-stealing, career-defining performance and one of the greatest live concerts in rock history.

\vspace{1em}

\textbf{Generated excerpt:}
\begin{quote}\itshape
Queen featured ``Hammer to Fall'' in position 3 during Live Aid 1985. BBC retrospectives highlight the crowd’s electrifying call-and-response led by Mercury and the performance’s enduring reputation as Live Aid’s pinnacle. CNN framed Queen’s Live Aid set as the moment they `stole the show', citing precision, charisma, and unmatched audience connection. Rolling Stone's review stated Queen never sounded better and many critics said they stole the show, even though U2 put up a very good fight. Wikipedia's review stated the Live Aid article’s Legacy section frequently cites Queen’s set as one of the greatest live performances in rock history and a defining moment of the concert. DiscoverMusic’s article declared Queen’s Live Aid set the day’s show-stopping event and `one of the greatest live concerts of all time'.
\end{quote}

This segment is a good example of faithful behaviour. The model correctly binds the positional setlist fact (``Hammer to Fall'' in position~3) to Live Aid~1985, then threads all five review statements into a coherent paragraph without altering their meaning. The wording remains close to the KG text while still reading fluently, and the passage foregrounds sources and evaluations in a way that fits Luca’s preference for detailed, review-driven context. No factual distortions or hallucinated claims are introduced.


\subsection{Positive deviation (Beat~0: introduction)}

\textbf{Retrieved KG facts.}  
The KG provides a descriptive paragraph about Live Aid~1985: a dual-venue global benefit concert held on 13~July~1985, inspired by the Band Aid charity single \emph{``Do They Know It’s Christmas?''} to raise funds for famine relief in Ethiopia, conceived by Bob~Geldof and Midge~Ure, broadcast to nearly two billion viewers worldwide, and described as a landmark in humanitarian fundraising and popular culture.


\vspace{1em}

\textbf{Generated excerpt:}
\begin{quote}\itshape
Live Aid 1985 was a dual-venue global benefit concert held on 13~July~1985. It was inspired by the Band Aid charity single ``Do They Know It’s Christmas?'' to raise funds for famine relief in Ethiopia. Bob Geldof and Midge Ure conceived Live Aid 1985. The concert was broadcast to nearly 2~billion viewers worldwide, and Live Aid 1985 became a landmark in humanitarian fundraising and popular culture.
\end{quote}

The factual content here matches the KG description exactly, but the model improves the structure by segmenting a long KG sentence into shorter clauses and by making the causal link between global reach and legacy explicit. Phrases such as \emph{``The concert was broadcast to nearly 2~billion viewers worldwide, and Live Aid 1985 became a landmark\dots''} are not present verbatim in the KG but constitute a useful, evidence-aligned inference. This is a positive deviation: the model enhances readability and causal flow without modifying or inventing facts.


\subsection{Negative deviation (Beat~2: performance detail)}

\textbf{Retrieved KG facts.}  
Separate descriptions of iconic performances, including: (a) a career-defining set in Philadelphia involving jokes about ``keeping the jacket on'', and (b) a different artist who performed at both venues on the same day by travelling via Concorde.

\vspace{1em}

\textbf{Generated excerpt:}
\begin{quote}\itshape
Additionally, by artists not specified in the context, the event included a career-defining set in Philadelphia, which featured quips about keeping the jacket on and then delivered a hit-packed performance, and the only artist to play both Wembley and Philadelphia the same day by Concorde, who also drums for Clapton and Led Zeppelin.
\end{quote}

Here, the system mistakenly merges two distinct KG rows into a single sentence. The performer remains unnamed despite being identifiable in the underlying data, and the fused structure introduces syntactic irregularities (e.g., \emph{``performance, and''}). This demonstrates a limitation of KG-only conditioning: when entries are sparse or loosely structured, the model may collapse them into a single, ambiguous narrative fragment.


\subsection{Additional noteworthy behaviours}\label{ssec:behaviours}

Beyond the focal examples above, several patterns recurred across multiple KG--RAG runs. We group them as follows, ordered by severity.

\begin{itemize}
    \item \textbf{Fact fusion and misbinding.}  
    In some runs, the model blended distinct KG rows into a single sentence, as in the iconic performance example where a career-defining Philadelphia set and a separate Concorde-enabled cross-venue appearance were implicitly treated as one continuous story. While each fragment is grounded in evidence, their conflation can mislead readers about performer identities or event structure.

    \item \textbf{Irrelevant tail facts from mixed evidence.}  
    When a beat combined heterogeneous KG rows (e.g., Queen’s Live Aid reviews plus an unrelated biographical fact about Madonna), the model occasionally appended a trailing sentence that surfaced the extra fact (such as Madonna’s birth date and birthplace). These additions are still factually correct but weaken topical focus and illustrate how broad evidence windows can leak marginal content into otherwise coherent beats.

    \item \textbf{Alias leakage from SPARQL queries.}  
    Some outputs surfaced the SPARQL query aliases directly (e.g.,
    \emph{``KG's eventName was Live Aid 1985''} or \emph{``Live Aid 1985's
    locationName was John F. Kennedy Stadium''}). These aliases (\texttt{eventName},
    \texttt{locationName}, \texttt{allWorkName}) were introduced in the refined SPARQL queries to group related data and reduce repetition. While the factual grounding is preserved, the inclusion of these variable names degrades the text quality, yielding sentences that read as direct data transcriptions rather than curated storytelling.

    \item \textbf{Mechanical list rendering.}  
    Long performer or song lists were sometimes carried over almost verbatim from KG string fields, including punctuation artefacts (extra periods, broken line breaks, or camel-cased identifiers such as \emph{RogerTaylor}). Similarly to the previous behaviour, this preserves coverage but reduces readability and undermines the narrative.

    \item \textbf{Over-literal table-to-text templates.}  
    For schedule-like data and location triples, the model frequently adopted a rigid pattern (\emph{``Live Aid 1985 had a location named John F. Kennedy Stadium. John F. Kennedy Stadium was in the city named Philadelphia. Philadelphia was in the country named United States of America''}). These outputs are accurate but read as stepwise graph traversal rather than an integrated description, which is stylistically at odds with Luca’s more curated, analytical persona.


    \item \textbf{Redundancy and light repetition.}  
    Across runs, some beats occasionally repeated the same information with small syntactic variations (for example, restating that Live Aid became a landmark in humanitarian fundraising and popular culture, or repeating founder names). These repetitions do not compromise factual accuracy but slightly inflate length relative to the target word range for Luca.
\end{itemize}

Jointly, these behaviours suggest that the KG--RAG pipeline is generally factually reliable at the sentence level, but sensitive to how KG rows are grouped and presented to the model.
The most impactful errors arise when distinct rows are fused or when heterogeneous evidence is mixed within a single beat.
One the other hand, minor issues concern style, schema exposure and redundancy.

\subsection{CQ-query template examples}\label{ssec:cq-query-templates}
We provide examples of the parameterised SPARQL templates mapped to the competency questions (CQs).
These templates serve as the explicit API contract used during the retrieval stage to gather targeted evidence from the KG.

Note that the \texttt{ex} namespace used in these templates refers to a compact ontology extension developed specifically for the Live Aid KG to simplify the querying process (\textit{e.g.}, by introducing dedicated binary relationships).

\vspace{.5em}

\begin{lstlisting}[language=SPARQL]
#CQ-ID:CQ-L14 Who were the founding members of [Artist]?

PREFIX schema: <http://schema.org/>
PREFIX mm: <https://w3id.org/polifonia/ontology/music-meta/>
PREFIX ex: <http://wembrewind.live/ex#>
SELECT ?artistName (GROUP_CONCAT(DISTINCT ?founderName; separator=", ") AS ?foundingMembers)
WHERE {
    BIND ({musicgroup} AS ?artist)
    ?artist a mm:MusicEnsemble ;
            schema:name ?artistName ;
            schema:foundingDate ?foundingYear .
    ?role a  mm:MusicEnsembleMembership ;
          mm:involvesMusicEnsemble ?artist ;
          mm:involvesMemberOfMusicEnsemble ?member ;
          ex:isOriginalMember true .
    ?member schema:name ?founderName .
}
GROUP BY ?artistName
\end{lstlisting}

\begin{lstlisting}[language=SPARQL]
#CQ-ID:CQ-E3 Which [Artist] performed at [Event] held at [Venue]?

PREFIX ex: <http://wembrewind.live/ex#>
PREFIX schema: <http://schema.org/>
PREFIX oa: <http://www.w3.org/ns/oa#>
PREFIX dct: <http://purl.org/dc/terms/>
SELECT ?sourceUrl ?exactSelector ?refinedBy
WHERE {
    BIND ({event} AS ?event)
    BIND ({venue} AS ?venue)
    ?annotation a oa:Annotation ;
           dct:subject ex:Performances ;
           schema:spatial ?venue ;
           oa:hasTarget ?event ;
           oa:hasBody ?background .
    ?background oa:hasSource ?sourceUrl ;
                oa:hasSelector ?secSel .
    OPTIONAL {
        ?secSel oa:refinedBy ?subSel .
        ?secSel oa:exact ?exactSelector .
        OPTIONAL { ?subSel oa:exact ?refinedBy }
        OPTIONAL { ?subSel oa:start ?start ; oa:end ?end }
      }
    ?event schema:name ?eventName .
}
\end{lstlisting}

\begin{lstlisting}[language=SPARQL]
#CQ-ID:CQ-L10 What was [Artist]'s full setlist at [Event]?

PREFIX ex: <http://wembrewind.live/ex#>
PREFIX mm: <https://w3id.org/polifonia/ontology/music-meta/>
PREFIX schema: <http://schema.org/>

SELECT ?eventName ?artistName ?song ?position
WHERE {
    BIND ({event} AS ?event)
    BIND ({musicgroup} AS ?artist)
  ?performance a mm:LivePerformance ;
               schema:performer   ?artist ;
               schema:isPartOf   ?event ;
               ex:setList ?setlist .
  ?setlist schema:itemListElement ?listItem .
  ?listItem schema:item ?work .
  ?listItem schema:position ?position .
  ?work schema:name ?song .
  ?artist schema:name ?artistName .
  ?event schema:name ?eventName .
}
ORDER BY ?position
\end{lstlisting}

\end{document}